\documentclass[letterpaper]{article} % DO NOT CHANGE THIS
\usepackage{aaai24}  % DO NOT CHANGE THIS
\usepackage{times}  % DO NOT CHANGE THIS
\usepackage{helvet}  % DO NOT CHANGE THIS
\usepackage{courier}  % DO NOT CHANGE THIS
\usepackage[hyphens]{url}  % DO NOT CHANGE THIS
\usepackage{graphicx} % DO NOT CHANGE THIS
\usepackage{natbib}  % DO NOT CHANGE THIS AND DO NOT ADD ANY OPTIONS TO IT
\usepackage{caption} % DO NOT CHANGE THIS AND DO NOT ADD ANY OPTIONS TO IT
\usepackage{algorithm}
\usepackage{algorithmic}
\usepackage{graphicx}

\usepackage{amsmath}
\usepackage{amsfonts}
\usepackage{subfigure}
\usepackage{color}
\usepackage{multirow}

\usepackage{newfloat}
\usepackage{listings}
\DeclareCaptionStyle{ruled}{labelfont=normalfont,labelsep=colon,strut=off} % DO NOT CHANGE THIS
\floatstyle{ruled}
\newfloat{listing}{tb}{lst}{}
\floatname{listing}{Listing}
\title{One-Step Forward and Backtrack: Overcoming Zig-Zagging in Loss-Aware Quantization Training}
\author{
    Lianbo Ma\textsuperscript{\rm 1}, Yuee Zhou\textsuperscript{\rm 1}, Jianlun Ma\textsuperscript{\rm 1}, Guo Yu\textsuperscript{\rm 2}\thanks{Corresponding author.}, Qing Li\textsuperscript{\rm 3}\\
}
\affiliations{
     \textsuperscript{\rm 1}Software College, Northeastern University, Shenyang, China\\
	\textsuperscript{\rm 2}Institute of Intelligent Manufacturing, NanJing Tech University, Nanjing, China\\
	\textsuperscript{\rm 3}Peng Cheng Laboratory, Shenzhen, China\\
    malb@swc.neu.edu.cn, \{zhouyuee, jianlunma\}@stumail.neu.edu.cn, guo.yu@njtech.edu.cn, liq@pcl.ac.cn

}

\usepackage{bibentry}
\begin{document}

\maketitle

\begin{abstract}
Weight quantization is an effective technique to compress deep neural networks for their deployment on edge devices with limited resources. Traditional loss-aware quantization methods commonly use the quantized gradient to replace the full-precision gradient. However, we discover that the gradient error will lead to an unexpected zig-zagging-like issue in the gradient descent learning procedures, where the gradient directions rapidly oscillate or zig-zag, and such issue seriously slows down the model convergence. Accordingly, this paper proposes a one-step forward and backtrack way for loss-aware quantization to get more accurate and stable gradient direction to defy this issue. During the gradient descent learning, a one-step forward search is designed to find the trial gradient of the next-step, which is adopted to adjust the gradient of current step towards the direction of fast convergence. After that, we backtrack the current step to update the full-precision and quantized weights through the current-step gradient and the trial gradient. A series of theoretical analysis and experiments on benchmark deep models have demonstrated the effectiveness and competitiveness of the proposed method, and our method especially outperforms others on the convergence performance.
\end{abstract}

\section{Introduction}
With the increase of computation and storage consumption in various cloud and edge computing applications, it is challenging to deploy deep neural networks (DNNs) on computation-constrained devices \cite{1r, 2r}. To enhance the efficiency of DNNs (e.g., low-power inference), it is critical to compress deep models with comparable performance \cite{3r, 4r}. Typical compression approaches include low-rank decomposition \cite{5r, 6r}, model pruning \cite{7r, 8r, 9r, 10r, 11r}, knowledge distillation \cite{12r, 13r, 14r, 39r}, and low-bit quantization \cite{15r, 16r, 17r, 18r, 19r}.

In this paper, we focus on the compression approaches of low-bit quantization, which are to reduce the model size with no or minor performance degradation. Specifically, the low-bit quantization is to quantize full-precision weights (or activations) to low bitwidth ones \cite{19r, 20r}. To this end, pioneer quantization approaches strive to find the closest low-precision approximation of the full-precision weights during the model learning process \cite{21r, 22r}. However, these approaches neglect the effect of quantization on the final loss, so that they may suffer from the severe accuracy degradation problem \cite{23r, 24r}. This problem becomes even more serious if the extremely low bitwidth fixed point (e.g., 1-bit) representation is used in the quantization \cite{25r}. Therefore, many studies on loss-aware quantization (LAQ) have proposed to solve the above issue \cite{26r,27r,28r,29r,30r}, and the main idea of LAQ is to optimize the quantized weights rather than the full-precision ones during the minimization of the loss function.

\begin{figure}[t] \centering
\subfigure[LAQ] {\label{fig.5.(a)}
\includegraphics[width=0.47\linewidth]{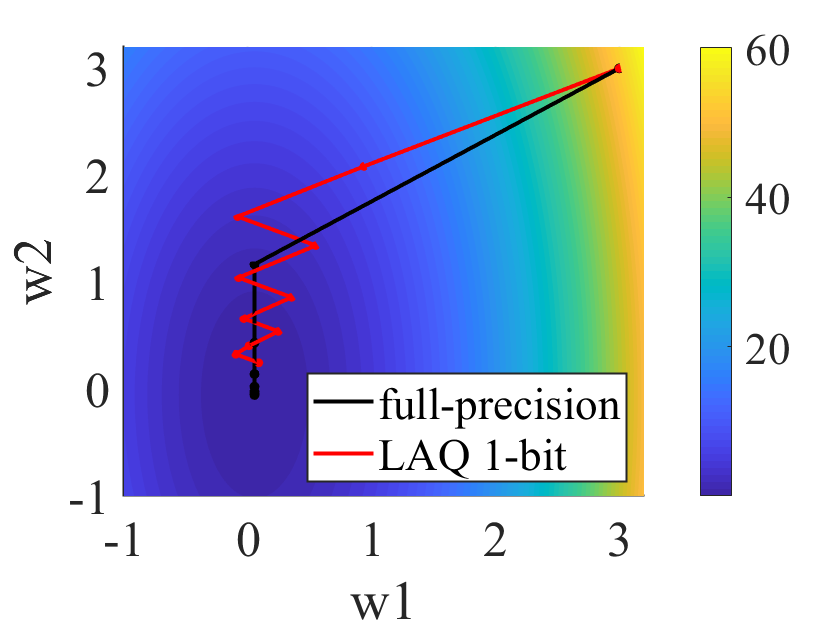}
}
\subfigure[BLAQ] {\label{fig.5.(b)}
\includegraphics[width=0.47\linewidth]{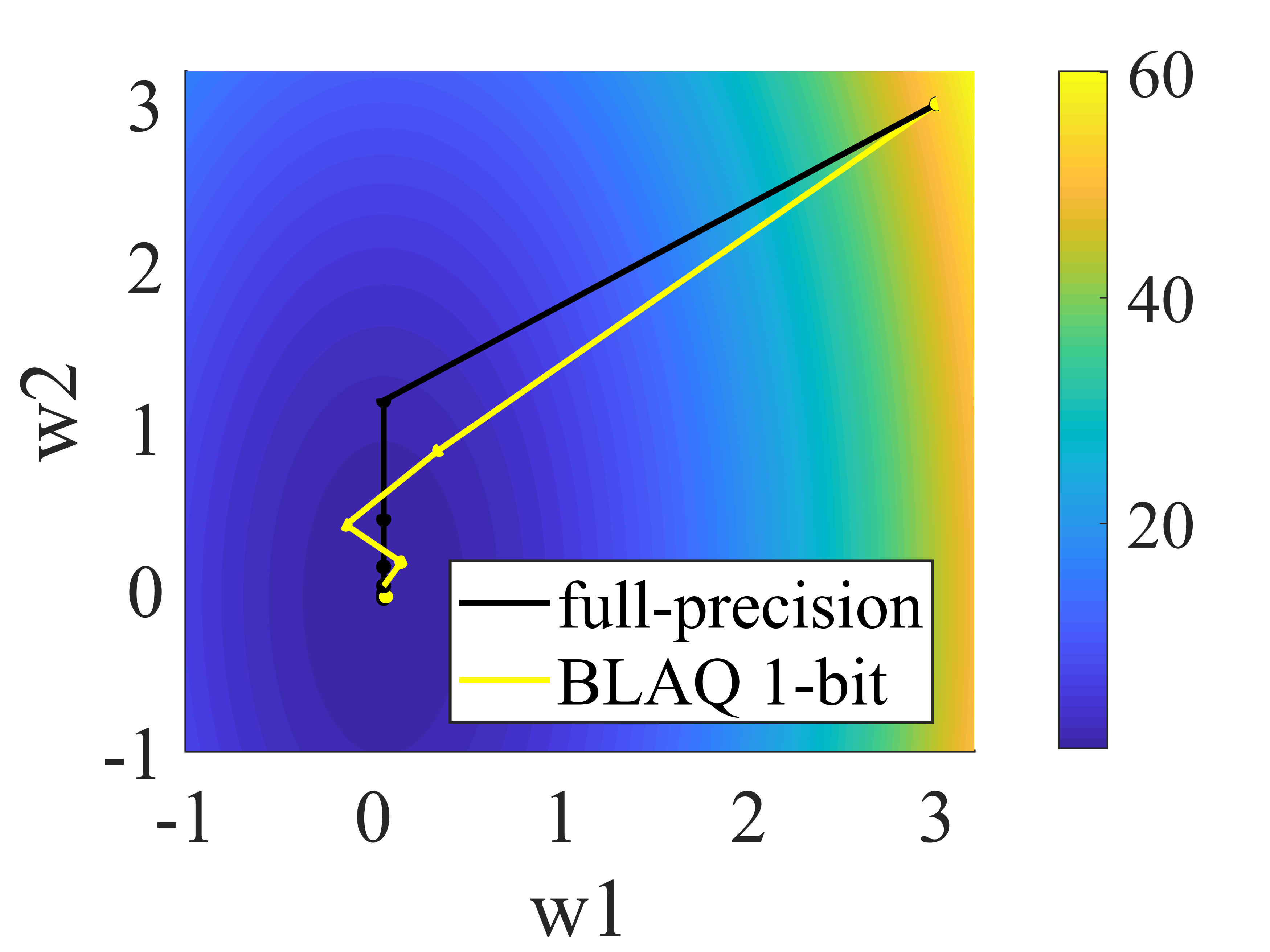}
}
\caption{Trajectories of weight updating obtained by LAQ and BLAQ in 1-bit quantization on the loss function $\ell(\omega)=5\left(\omega_{1}-0.054\right)^{2}+\left(\omega_{2}+0.055\right)^{2}$. 
The black line shows the evolution of full-precision weight values via iterations during training, while the red and yellow lines are the evolution of 1-bit quantization case.
}
\label{fig1}
\end{figure}

However, we discover that existing LAQ methods \cite{27r,28r} may suffer from the zig-zagging-like issue that the gradient directions rapidly oscillate or zig-zag during the gradient descent learning procedures, and such issue seriously slows down the model convergence.
An example is shown in Fig.\ref{fig1} (a) that the quantized gradient direction at each step tends to be zig-zag or rapidly oscillated and the search with quantized weights needs more epochs to get converged in comparison with the search with full-precision weights. In some extreme cases, such issue even makes the search fail to converge and end up with oscillation.  Unfortunately, 
most recently published studies do not realize this issue but compensates for the gradient quantization error by enhancing quantization representations\cite{17r,21r,28r,29r}. Due to the big gap between extremely low-bit (e.g.,1-bit) representations and full-precision (e.g., 32-bit) ones, it is hard to make full compensation of quantization error.

Inspired by the numerical analysis theory \cite{30r} that iteratively using the trial results backtracked from next-step search to update next-step items can contribute to the numerical stability, we try to tackle the above zig-zagging-like issue in a different way, i.e., improving the quantization updating rules to get more accurate and stable gradient direction. In other words, the key to reduce the zig-zagging issue in this study is to explore trial information from the next-step exploratory search and take them as extra gradient compensation to enhance the stability of weight updating.

Following the above idea, we propose an effective backtracking-search loss-aware quantization (BLAQ) method, which optimizes weights in a one-step forward and backtrack way to get more accurate and stable gradient direction to defy the zig-zagging-like issue. During the gradient descent learning, the one-step forward search is to find a trial gradient of the next step to drive the gradient of current step towards the direction of fast convergence. After that, 
we backtrack the current step to update the full-precision and quantized weights through the current-step gradient and the trial gradient. As a result, the estimation of gradient direction is more accurate and stable than the traditional methods\cite{17r,21r,28r,29r}. A pilot study is shown in Fig.\ref{fig1} (b) that BLAQ is able to effectively and efficiently reduce redundant zig-zagging steps and converged faster in comparison with LAQ. 

The main contributions of this paper include:
\begin{itemize}
\item We discover the zig-zagging-like issue in the search of quantized weights via LAQ, and find that the issue can seriously slow down the model convergence.
    	
\item  We propose a novel loss-aware quantization method to defy the zig-zagging-like issue, which generates low-bit quantized network in a one-step forward and backtrack way. Therefore, a new quantization framework with good convergence property is provided in this study.

\item We provide theoretical analysis to show that the proposed quantization is mathematically better than other counterparts in convergence properties, and the effectiveness of our method is also verified on a set of deep models.
\end{itemize}

\section{Preliminaries}

\subsection{Notations for Weight Quantization}

In this paper, we use $W$ to represent a full-precision neural network, and $\hat{W}$ to represent a quantization neural network. For a convolution neural network, its full-precision weights of all $L$ layers are defined as $W$=($W_{1},W_{2},\cdots,W_{L}$), where $\omega_{l}=vec(W_{l})$ and $W_{l}\in{R^{h \times h \times c_{i} \times c_{o}}}$ is the weight tensor at layer $l$. Here,  $h$, $c_{i}$ and $c_{o}$ are the size of convolution kernel, the numbers of input and output channels, respectively. The corresponding quantized weights are denoted by $\hat{W}$=(${\hat{W}_{1},\hat{W}_{2},\cdots,\hat{W}_{L}}$), where  ${\hat{\omega}}_{l}=vec({\hat{W}}_{l})$ and ${\hat{W}}_{l}$ is a quantization weight tensor at layer $l$. Here, $vec(X)$ outputs a vector by stacking the columns of a matrix $X$, and $diag(X)$ a diagonal matrix, where the diagonal elements are constructed by the diagonal of $X$. We refer to ${D}_{l}$ and ${\hat{D}}_{l}$ as the approximated diagonal Hessian matrix for ${W}_{l}$ and ${\hat{W}}_{l}$, respectively. $Q$ is a set of $2k$ quantized values, where $Q$ = $\{-1, \cdots, -\frac{2}{2^{k-1}}, -\frac{1}{2^{k-1}}, \frac{1}{2^{k-1}}, \frac{2}{2^{k-1}}, \cdots, 1\}$. %where $k=2^{m-1}+1$. There are two schemes for $Q$, including  $\{-1,-\frac{k-1}{k},\cdots,-\frac{1}{k},0,\frac{1}{k},\cdots,\frac{k-1}{k},1\}$ (linear quantization) and $\{-1,-\frac{1}{2},\cdots,-\frac{1}{2^{k-1}},0,\frac{1}{2^{k-1}},\cdots,\frac{1}{2},1\}$ (linear logarithmic). 
When $k=1$, $Q=\{-1,1\}$.
 ``$\oslash$'' is used as elementwise division and ``$\odot$'' as elementwise multiplication. $<x,y>$ is the inner product of vectors $x$ and $y$. Besides,  ${\Vert \omega \Vert^2_D}={\omega}^TD\omega$, and $[\omega]_Q$ denotes the rounding entries of $\omega$ to the closed fixed point in $Q$.

\subsection{Loss-Aware Weight Quantization}
In the forward propagation, LAQ \cite{27r} utilizes the second-order Taylor expansion to quantize the full-precision weights, and then uses the proximal Newton method to solve the optimization problem \cite{32r}.
At iteration $t$, the objective function of LAQ is replaced by the second-order series expanded at $\hat{\omega}^t=\alpha^t\beta^t$
\begin{equation}\label{Eq1}
\begin{aligned}
\min ~&\ell\left(\hat{\omega}^{t-1}\right)+\hat{g}^{T^{t-1}}\left(\hat{\omega}^{t}-\hat{\omega}^{t-1}\right)\\
& +\frac{1}{2}\left(\hat{\omega}^{t}-\hat{\omega}^{t-1}\right)^T \hat{H}^{t-1}\left(\hat{\omega}^{t}-\hat{\omega}^{t-1}\right), \\
s.t.~&{\hat{\omega}}^{t}=\alpha^{t}\beta^{t}, \alpha^{t}=\frac{\Vert\omega^{t}\Vert}{n}, \beta^{t}=sign\left(\omega^{t}\right),
\end{aligned}
\end{equation}
\noindent where $\hat{g}^{t-1}$ and $\hat{H}^{t-1}$ are the first and second derivatives of the loss function $\ell(\hat{\omega}^{t-1})$ with respect to quantized weight ${\omega}^{t-1}$, respectively. 

For DNNs, the Hessian matrix $H^{t-1}$ is rarely positive semi-definite and intractable to calculate. Hence, the diagonal Hessian matrix $D^{t-1}$ is suggested to approximate $H^{t-1}$ (i.e., $D=diag(H)$) \cite{33r}. Therefore, Eq. \ref{Eq1} can be rewritten as (The derivation process is provided in the Section 2 of the Appendix \footnote{You can refer to our appendix on the following website: https://github.com/paperProof24/Appendix\_BLAQ}.)
\begin{equation}\label{Eq2}
\begin{aligned}
{\arg\min}_{\hat{\omega}^t}\frac{1}{2}\Vert {\omega}^t-\hat{\omega}^t \Vert^2_{{\hat{D}}^{t-1}},
\end{aligned}
\end{equation}
\noindent where
\begin{equation}\label{Eq3}
\begin{aligned}
{\omega}^t=\hat{\omega}^{t-1}-\hat{g}^{t-1}\oslash\hat{D}^{t-1}.
\end{aligned}
\end{equation}
Eq. \ref{Eq2} and Eq. \ref{Eq3} are used to update the quantized and full-precision weights, respectively.
\subsection{Zig-zagging-like Issue of Weight Updating}
We discover that the zig-zagging-like issue easily occurs in LAQ methods that the gradient directions rapidly oscillate or zig-zag during the gradient descent learning procedures
and the gradient quantization error seriously slow down the model convergence, especially for extremely low bit-width quantization. A pilot experiment is given in Fig. \ref{fig1} (a) that the gradient direction is sharply oscillated and evolution takes many steps to get converged.

Theoretically, the full-precision weights $\omega^t$ and quantized weights $\hat{\omega}^t$ in Eqs. \ref{Eq2}-\ref{Eq3} are alternatively and iteratively computed. Note that ${\hat{g}}^{t-1}$ related to the quantized gradient is used to update $\omega^t$ in Eq. \ref{Eq3}. However, the loss of information from the low-bit quantization will make the quantization error about $\hat{g}^{t-1}$ unavoidable accumulated over the iterations. Once the accumulated error reaches a threshold, the zig-zagging-like phenomenon\footnote{In fact, the landscape feature of optimization problem is also a factor to generate zig-zagging phenomenon but this work only focuses on the gradient errors.} will occur. Even worse, Eqs. \ref{Eq2}-\ref{Eq3} may make the model fail to converge.
Given a simple toy loss function $f(\omega)=c\omega^\frac{3}{2}$, if Eqs. \ref{Eq2}-\ref{Eq3} are used to calculate quantized weights, for any initial weights $\omega_0$, the optimization of $f(\omega)$ cannot converge and end in oscillating. The proof is provided in the Section 3 of the Appendix.

%Notably, the zig-zagging phenomenon in our study is essentially different from the flip-flop in AdamBNN~\cite{48r} and oscillation~\cite{47r}. The flip-flop in~\cite{48r} refers to unstable training caused by the weights whose signs are changed after updating. The oscillation in~\cite{47r} refers to the problem that the straight-through estimator (STE) has latent weights oscillating around the decision boundary between adjacent quantized states. In contrast, the zig-zagging phenomenon depicts that the direction of updating weights will be deviated away from the optimum due to the quantization error of gradient.
Notably, the zig-zagging phenomenon in our study is essentially different from the oscillation~\cite{47r}. The zig-zagging-like issue mentioned in our work is essentially caused by the zig-zag fluctuation of the weights due to errors in the quantization of the gradient during training. Due to the inaccuracy of the quantized gradient, the weights are updated in such a way that they deviate from the original trajectory during training using the quantized gradient, leading to the zig-zagging-like issue. And the cause of the oscillation problem proposed by \cite{47r} is the inaccurate calculation of full-precision weights due to truncation error and rounding error when the full-precision weights are close to the quantization decision boundaries, leading to inaccurate quantization results and recurrent oscillations of the quantization weights. The external manifestations of the two oscillation phenomena may be somewhat similar, but the internal causes are different, so different solutions need to be taken. 

\subsection{Zig-zagging-like Issue in Practice}
\begin{figure}\centering
\subfigure[] { 
\includegraphics[width=0.45\columnwidth]{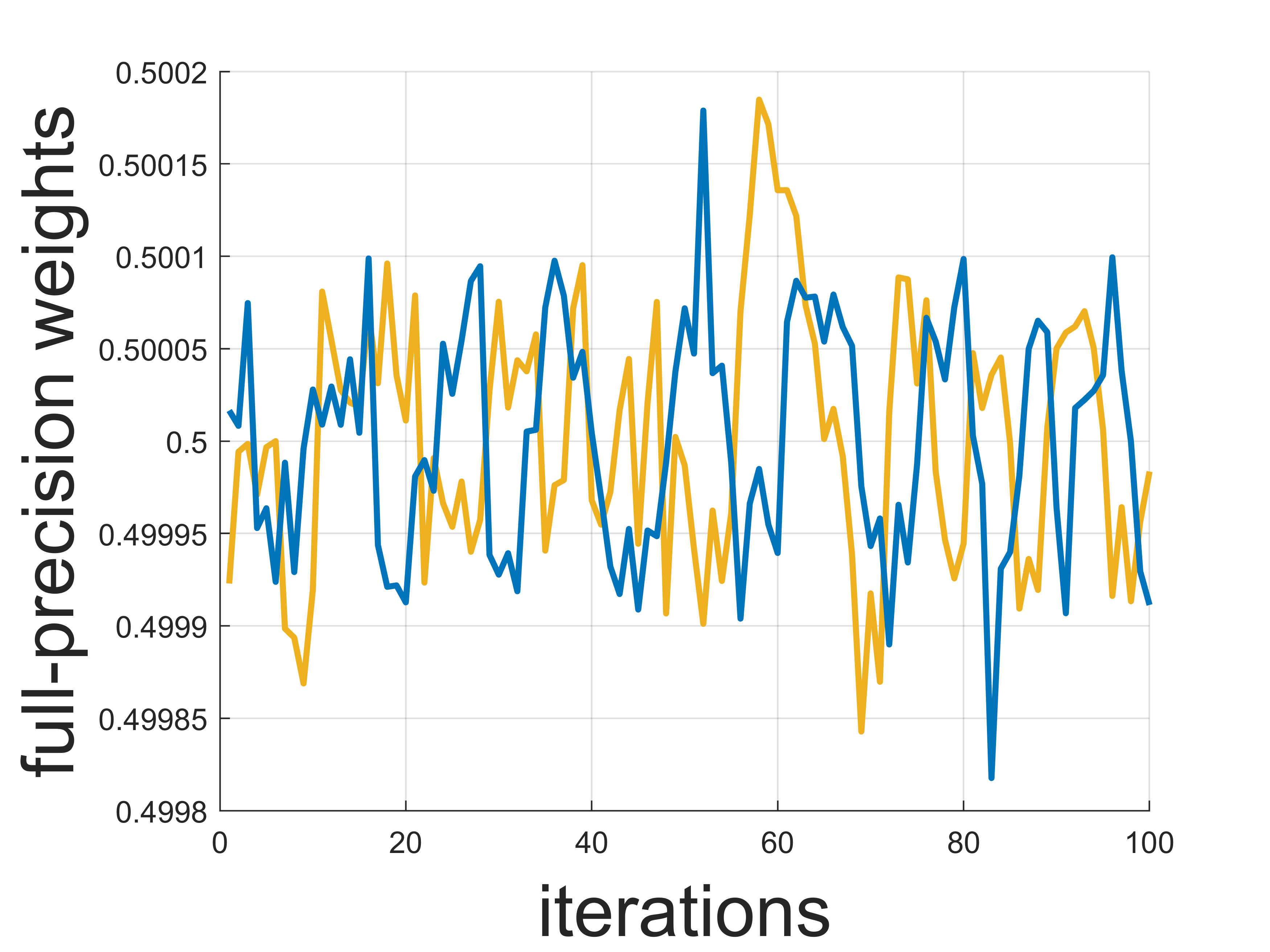}
}
\subfigure[] { 
\includegraphics[width=0.45\columnwidth]{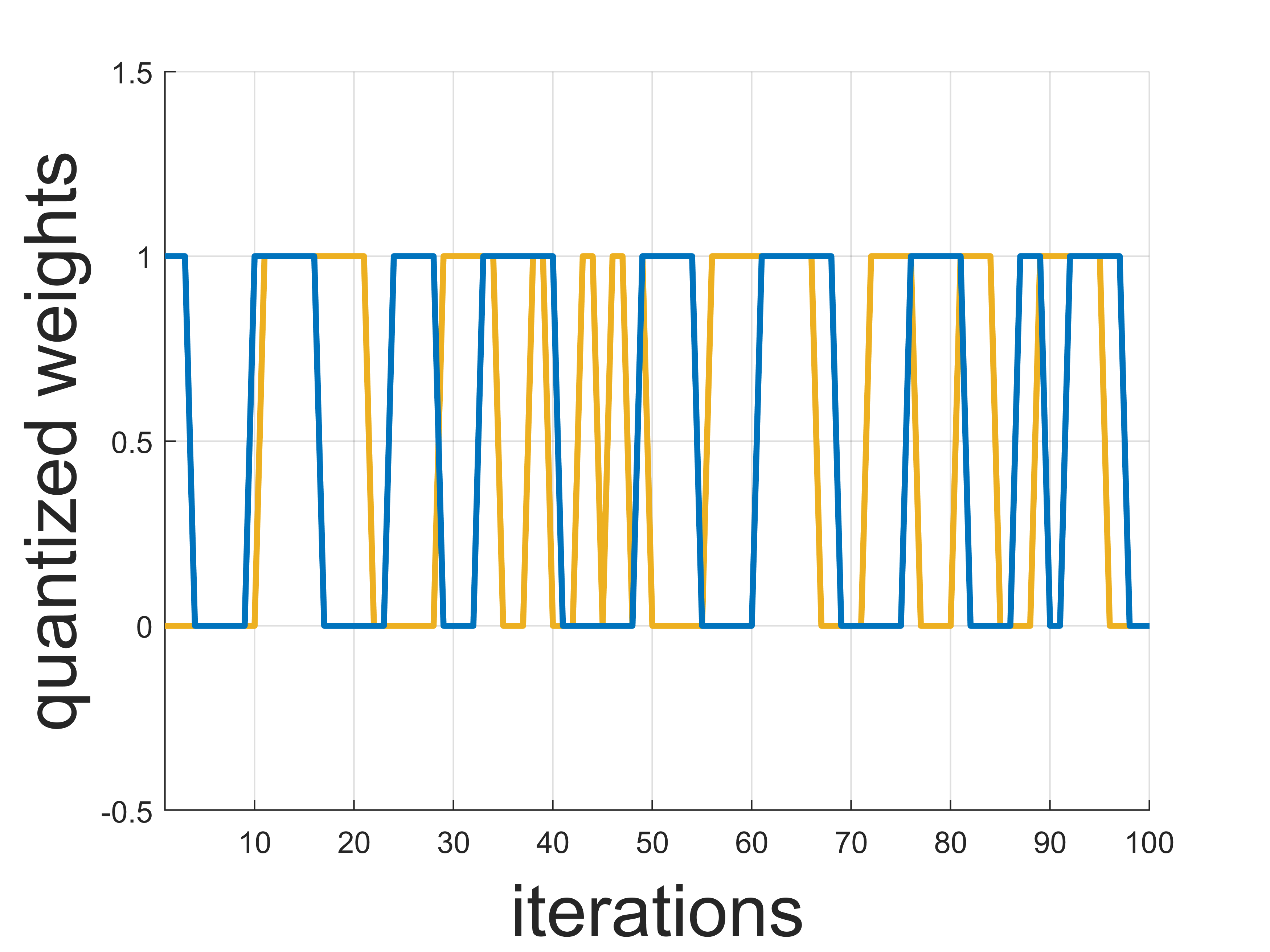}
}
\caption{Regression of randomly selected weights during the last 100 epochs of training ResNet18 on ILSVRC12 (ImageNet) during LAQ 2-bit quantization. The (a) show the full-precision weights, and the (b) plot the corresponding quantized weights.}\label{figVGG1}
\end{figure}

\begin{figure} \centering
\subfigure[] { 
\includegraphics[width=0.45\columnwidth]{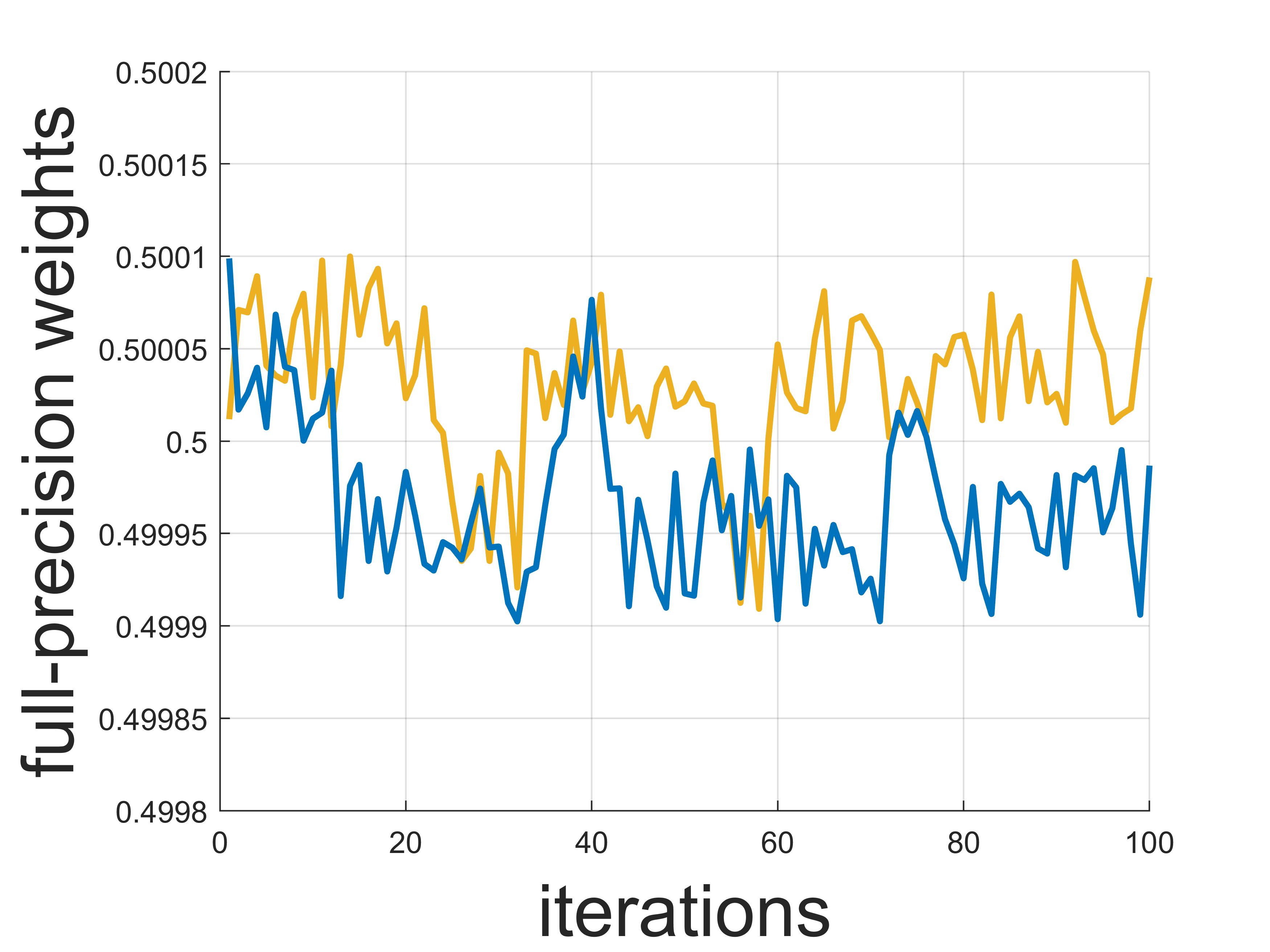}
}
\subfigure[] { 
\includegraphics[width=0.45\columnwidth]{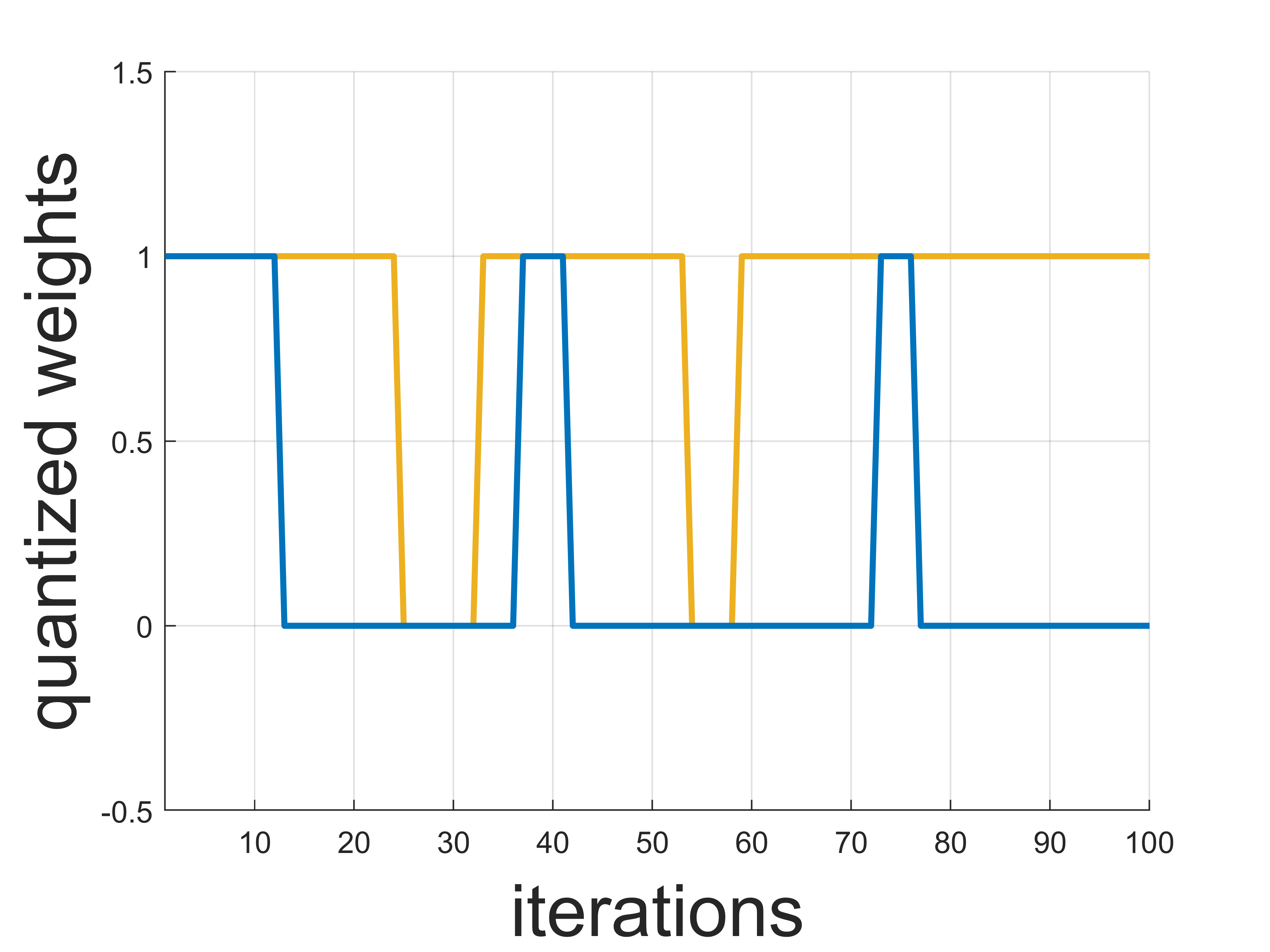}
}
\caption{Regression with the same weights as Fig. \ref{figVGG1} selected during the last 100 epochs of training ResNet18 on ILSVRC12 (ImageNet) during BLAQ 2-bit quantization. The (a) show the full-precision weights, and the (b) plot the corresponding quantized weights.}\label{figVGG2}
\end{figure}

The zig-zagging-like issue exists not only in the toy example presented in Fig. \ref{fig1}, but also in the quantization of large neural networks. An example is shown in Fig. \ref{figVGG1} that a clear zig-zagging-like phenomenon during the quantization of ResNet18 on ImageNet using the LAQ method. In addition, as shown in Fig. \ref{figVGG2}, the zig-zagging-like phenomenon is significantly improved during the quantization of ResNet18 on ImageNet using our proposed BLAQ method.
%An example is shown in Fig. \ref{figVGG1} that a clear zig-zagging-like phenomenon appears in plot \ref{figVGG1} (b) during the quantization. Besides, we can also observe the oscillation still exists after 70 iterations in plot \ref{figVGG2} (b), when the model should be converged.
\subsection{Convergence Result Analysis}
We also find that the lower the bitwidth of the quantization uses, the more intense the zig-zagging phenomenon of weights will be. The corresponding experimental results are presented in Table \ref{tablen} which shows that the maximum oscillation amplitude of the full-precision weights and oscillation frequencies of the quantized weights of the last 150 epochs during the training of ResNet18 on ILSVRC12 (ImageNet) using different quantization bitwidths. We can observe that when we use the lower bitwidth for quantization, the maximum oscillation amplitude of the full-precision weight becomes more intense, and the oscillation on the quantization weight becomes more frequent. In summary, the zig-zagging phenomenon is an important issue in the quantization, which can slow down or even prohibit the convergence of the model, especially in the case of extremely low-bit quantization.

\begin{table}[t]
	\center
	\begin{tabular}{cccc}
		\hline
		\textbf{Method}  &  bitwidth    &  amplitude &  frequency \\ \hline
		\multirow{3}*{LAQ}   & 1-bit     & 0.000486   &  15 \\ 
		         ~           & 2-bit     & 0.000342   &  11 \\ 
		         ~           & 4-bit     & 0.000261   &  8 \\  \hline 
		\multirow{3}*{BLAQ}  & 1-bit     & 0.000143   &  4 \\ 
		~                    & 2-bit     & 0.000102   &  3\\
		~                    & 4-bit     & 0.000067   &  1\\ \hline
	\end{tabular}
	\caption{The maximum oscillation amplitude of the full-precision weight and oscillation frequencies of the quantized weight for the same locked weight under the LAQ and BLAQ methods.}
	\label{tablen}
\end{table}

\section{Proposed Method}

\subsection{Loss-Aware Quantization via Two-stage Weight Updating}
In order to defy the above issue, we follow the procedure of LAQ \cite{27r} and come up with a novel one-step forward and backtrack quantization framework, which is terms as BLAQ for convenience. 

\begin{figure}[t]
	\centering
	\includegraphics[width=0.45\textwidth]{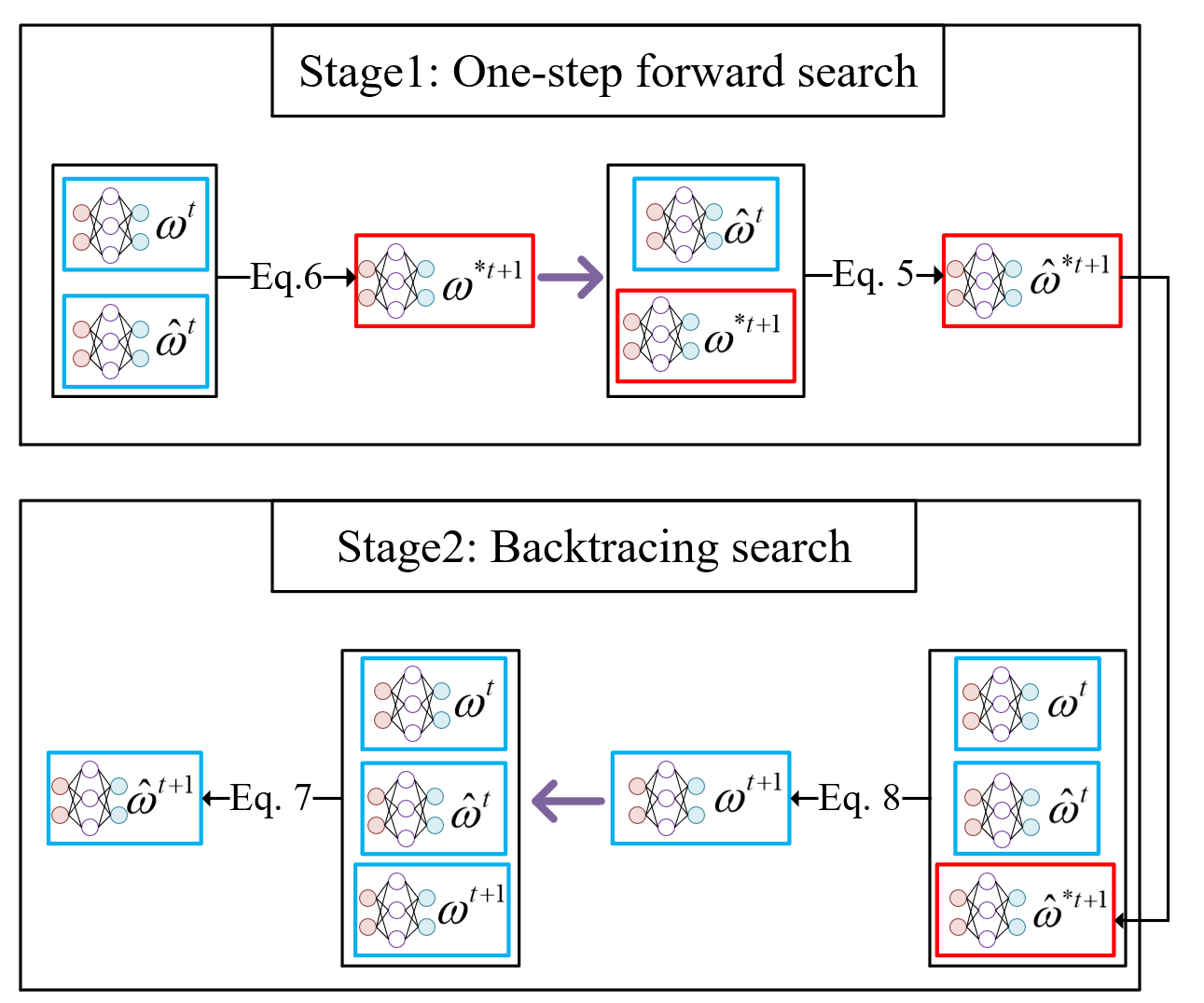}
	%文中引用该图片代号
	\caption{Two-stage updating process: the blue rectangles denote the real weights, and the red ones are the trial weights.}
	\label{fig2}
\end{figure}

The principal process of BLAQ is presented in Fig. \ref{fig2}.
In stage 1, the one-step forward search is performed. Given the real full-precision weights $\omega^t$ and quantized weights ${\hat{\omega}}^t$, we firstly use Eq. \ref{Eq6} to obtain the trial full-precision weights $\omega^{*(t+1)}$. Then, the real quantized weights and the trial full-precision weights are adopted to solve Eq. \ref{Eq5}, so that the trial quantized weights ${\hat{\omega}}^{*(t+1)}$ can be achieved. In stage 2, the backtrack quantization is executed. We firstly apply Eq. \ref{Eq8} to get the full-precision weights $\omega^{t+1}$ on the basis of the acquired weights from stage 1. After that, Eq. \ref{Eq7} is solved to obtain the real quantized weights ${\hat{\omega}}^{t+1}$ by using the weights from the previous steps. The details of the proposed BLAQ is introduced as follows. The corresponding pseudocode of BLAQ is presented in Algorithm 1 of the Appendix.

\textit{\textbf{Stage 1 (One-step forward search)}}: We take one forward step to tentatively obtain the trial quantized weights $\hat{\omega}^{*(t+1)}$. Therefore, the second-order Taylor expansion centered at quantized weights $\hat{\omega}^{*(t+1)}$ is applied, and the loss function is minimized by Eq. \ref{Eq4}. For convenience, we omit the subscript $l$ here:
\begin{equation}\label{Eq4}
\begin{aligned}
&{\arg\min}_{\hat{\omega}^{*(t+1)}}\ell(\omega^t)+{\hat{g}}^{tT}(\hat{\omega}^{*(t+1)}-\omega^{t})\\
&~~~~+\frac{1}{2}(\hat{\omega}^{*(t+1)}-\omega^{t})^T\hat{H}^{t}(\hat{\omega}^{*(t+1)}-\omega^{t}),\\
&s.t.~{\hat{\omega}}^{*(t+1)}=\alpha^{t+1}\beta^{t+1},\\
 &~~~~~\alpha^{t+1}>0, ~\beta^{t+1}=sign(\omega^{*(t+1)}),
\end{aligned}
\end{equation}
\noindent where $\hat{g}^t$ and $\hat{H}^t$ are the first and second derivations of the loss function $\ell(\hat{\omega}^t)$ with respect to the $\hat{\omega}^t$, respectively. Besides binarization, our method is also suitable for multi-bit quantization if we reset the value of the $\beta$ parameter in Eq. \ref{Eq4}. If the diagonal Hessian matrix $\hat{D}^t$ is used to approximate $\hat{H}^t$, Eq. \ref{Eq4} can be reformulated as
\begin{equation}\label{Eq5}
\begin{aligned}
{\arg\min}_{\hat{\omega}^{*(t+1)}}\frac{1}{2}\Vert {\omega}^{*(t+1)}-\hat{\omega}^{*(t+1)} \Vert^2_{\hat{D}^t},
\end{aligned}
\end{equation}
\noindent where
\begin{equation}\label{Eq6}
\begin{aligned}
\omega^{*(t+1)}=\omega^t-\hat{g}^t\oslash{\hat{D}^t}.
\end{aligned}
\end{equation}
Here $\alpha$ and $\beta$ are solved by using an alternating update approach, the corresponding pseudocode is presented in Algorithm 2 of the Appendix.

\textit{\textbf{Stage 2 (Backtracking search)}:} After stage 1, we will acquire the corresponding trial quantized weights $\hat{\omega}^{*(t+1)}$, trial quantized gradient $\hat{g}^{*(t+1)}$ and trial diagonal Hessian matrix $\hat{D}^{*(t+1)}$. Then, we backtrack to re-compute the real quantized weights $\hat{\omega}^{t+1}$ as follows:

Following the derivation of original LAQ (i.e., Eqs. \ref{Eq1}-\ref{Eq3}), the optimization formula at stage 2 can be inferred as
\begin{equation}\label{Eq7}
\begin{aligned}
{\arg\min}_{\hat{\omega}^{t+1}}\frac{1}{2}\Vert {\omega}^{t+1}-\hat{\omega}^{t+1} \Vert^2_{\hat{D}^{t+1}},
\end{aligned}
\end{equation}
\noindent where
\begin{equation}\label{Eq8}
\begin{aligned}
\omega^{t+1}=\omega^t-\hat{g}^{t+1}\oslash\hat{D}^{t+1}.
\end{aligned}
\end{equation}
After that, we will compute the real quantized gradients (i.e., $\hat{g}^{t+1}$ and $\hat{D}^{t+1}$) in Eq. \ref{Eq8} through the real quantized gradients (i.e., $\hat{g}^t$ and $\hat{D}^t$) at $(t)$th step, and the trial quantized gradients (i.e., $\hat{g}^{*(t+1)}$ and $\hat{D}^{*(t+1)}$) at $(t+1)$th step.
\begin{equation}\label{Eq9}
\begin{aligned}
\hat{g}^{t+1}=a\hat{g}^t+(1-a)\hat{g}^{*(t+1)},
\end{aligned}    
\end{equation}
\begin{equation}\label{Eq10}
\begin{aligned}
\hat{D}^{t+1}=a\hat{D}^t+(1-a)\hat{D}^{*(t+1)},
\end{aligned}      
\end{equation}
where $a$ is a coefficient and determined in the ablation study, (which is presented in the Section 6 of the Appendix). Such backtrack-search approach is able to get a more accurate estimation of the quantized gradient ${\hat{g}^{t+1}}$. 

According to Algorithm 1 of the Appendix, we can obtain the real quantized weights ${\hat{\omega}^{t+1}}$ by the proposed one-step forward and backtrack quantization framework, i.e., solving the Eqs. \ref{Eq7}-\ref{Eq8}. In Section 4 of the Appendix, we have also theoretically verified that BLAQ will not oscillate on the simple toy experiment, where the traditional LAQ method does.

\subsection{Convergence Analysis}
Following the common assumptions \cite{21r}, i.e., the loss function $\ell(\omega)$ is convex, twice differentiable, $L_1$-$smooth$ and  $\mu$-$strongly$ convex with respect to $\omega$, we give the following analysis. 

\textbf{Theorem 1.} For the loss function $\ell(\omega)$ with the learning rate $\eta^t$, the convergence of our method is as 
\begin{equation}\label{Eq11}
\begin{aligned}
\ell\left(\omega^{t+1}\right)-\ell\left(\omega^{*}\right) \leq \frac{L_{1}+L_{1}^{3}\left(\eta^{t+1}\right)^{2}-2 \mu^{2} \eta^{t+1}}{2} \Delta^{2}.
\end{aligned}     
\end{equation}

\textbf{Theorem 2.} When $\frac{2}{L_{1} \eta}-1<a<1$ holds, the convergence of our method is better than that of LAQ \cite{27r}.

The corresponding proof is presented in the Section 5 of the Appendix.

\section{Experiment}
\subsection{Experimental Setup}
Four widely-used datasets (CIFAR10, MNIST, SVHN and ILSVRC12 (ImageNet)) are applied to validate the performance of BLAQ. The baseline methods for comparison include: 1) BC \cite{34r}, 2) BWN \cite{19r}, 3) LAQ \cite{27r}, 4) TRQ \cite{35r}, 5) AQ \cite{36r}, 6) ALQ \cite{29r}, 7) DC \cite{16r}, 8) ADMM \cite{37r}, 9) LR \cite{49r}, 10) DSQ \cite{50r}, 11) TWN \cite{44r}, 12) LQ-Net \cite{45r}, 13) QIL \cite{51r}, and 14) OCTAV \cite{52r}. For deep network architectures, we conduct experiments with VGG \cite{27r} on CIFAR10, with LeNet5 \cite{29r} and 4-layer Model \cite{27r} on MNIST, with SVHNNet \cite{27r} on SVHN, and with ResNet18 \cite{29r} on ILSVRC12. See Section 7 of the Appendix for more details. In fact, although the optimal values of the hyperparameters $a$ and $m$ vary across different datasets and models, we conducted extensive ablation experiments and ultimately concluded that for small datasets and models, the optimal values of $a$ and $m$ are typically 0.6 and 5. For large datasets and models, the optimal values of $a$ and $m$ are usually 0.9 and 10. See Section 7 of the Appendix for more details. Therefore, in BLAQ, we set the hyperparameters $a$ and $m$ to 0.6 and 5 in the datasets CIFAR10, MNIST, and SVHN, and set the hyperparameters $a$ and $m$ to 0.9 and 10 in the dataset ILSVRC12.

\subsection{Effectiveness of BLAQ}
\begin{figure*}[htbp] \centering
\subfigure[VGG on CIFAR10] { \label{4.2.1}
\includegraphics[width=0.5\columnwidth]{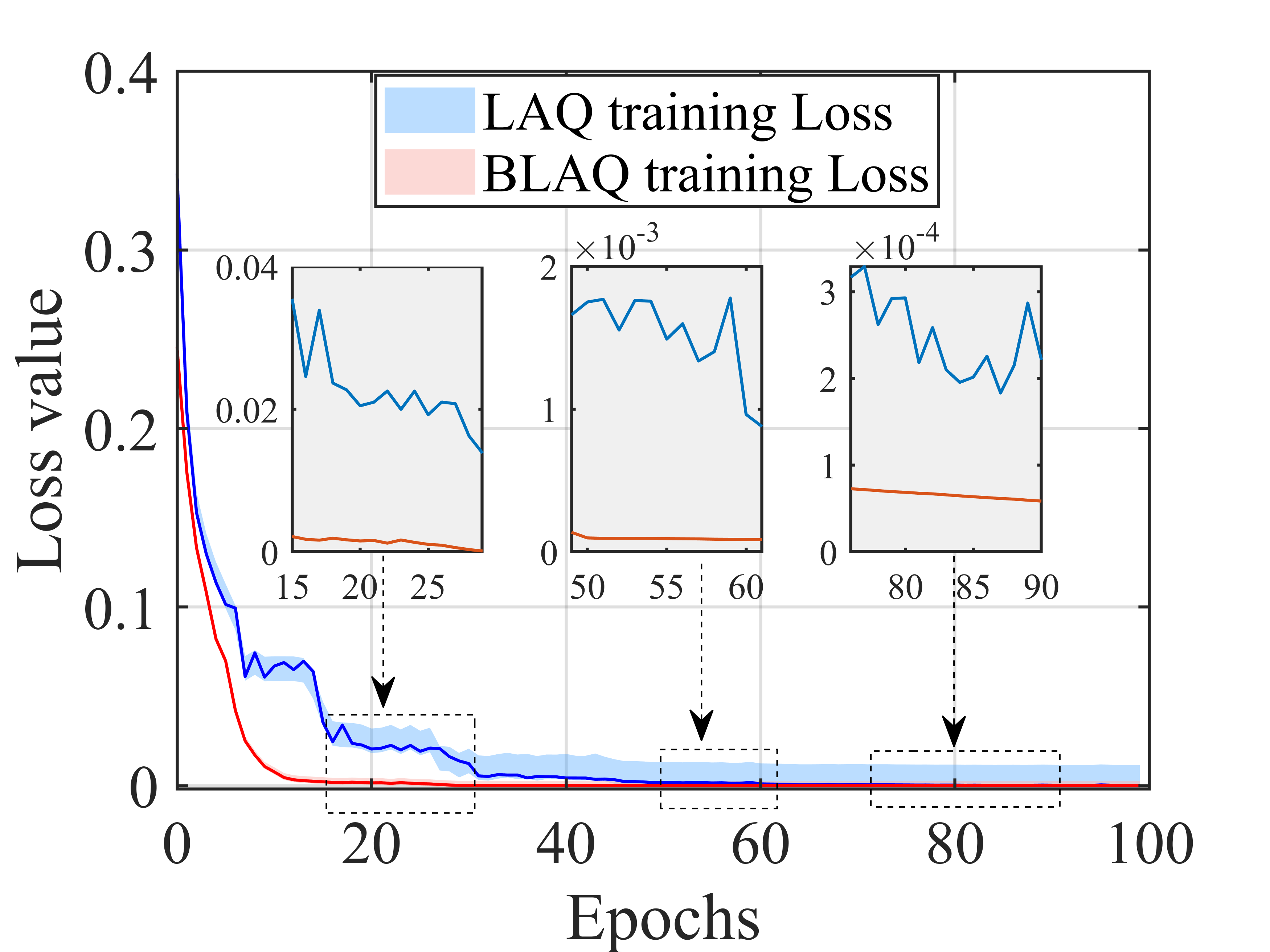}
}
\subfigure[4-layer Model on MNIST] { \label{4.2.2}
\includegraphics[width=0.5\columnwidth]{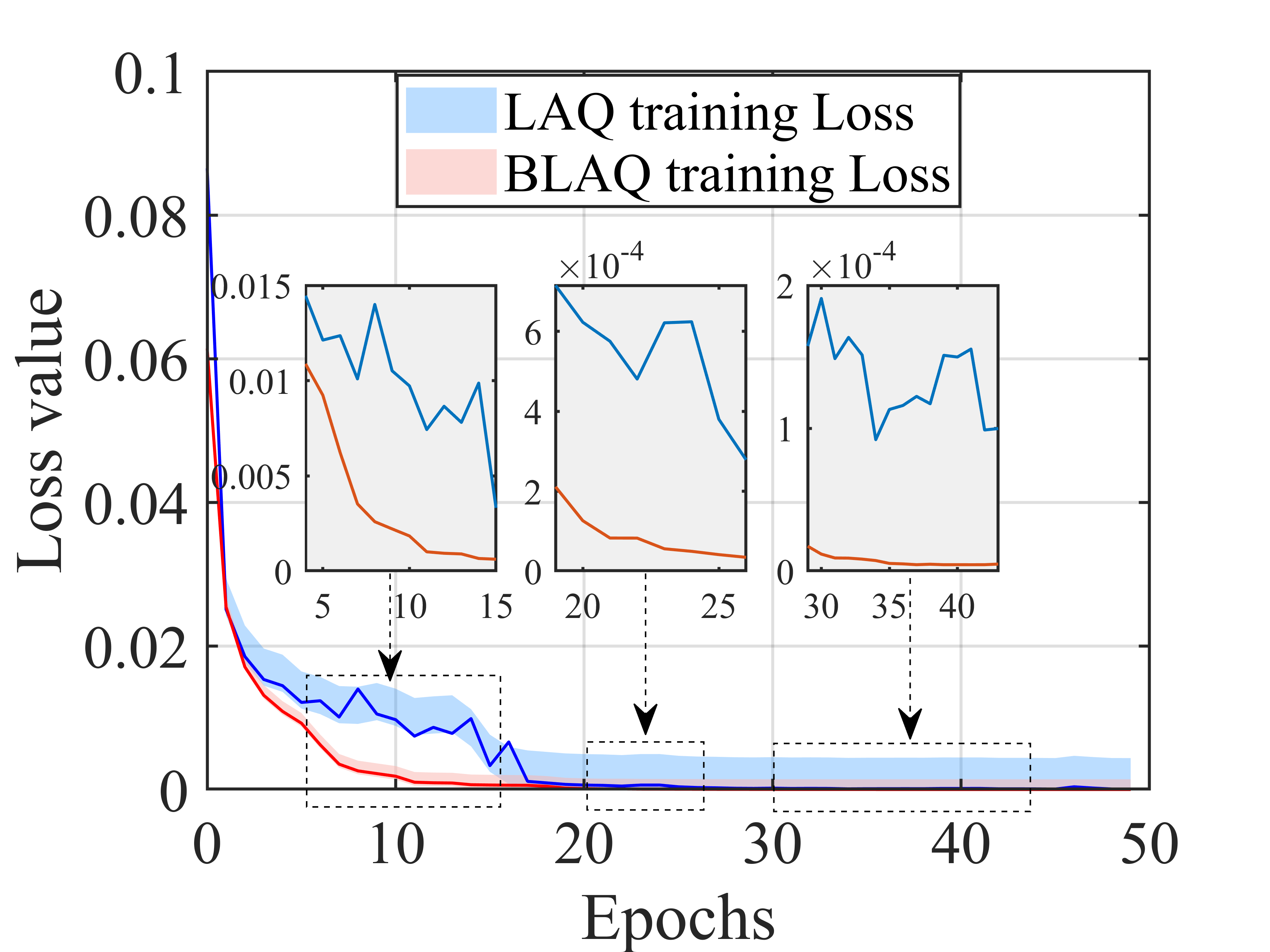}
}
\subfigure[SVHNNet on SVHN] { \label{4.2.3}
\includegraphics[width=0.5\columnwidth]{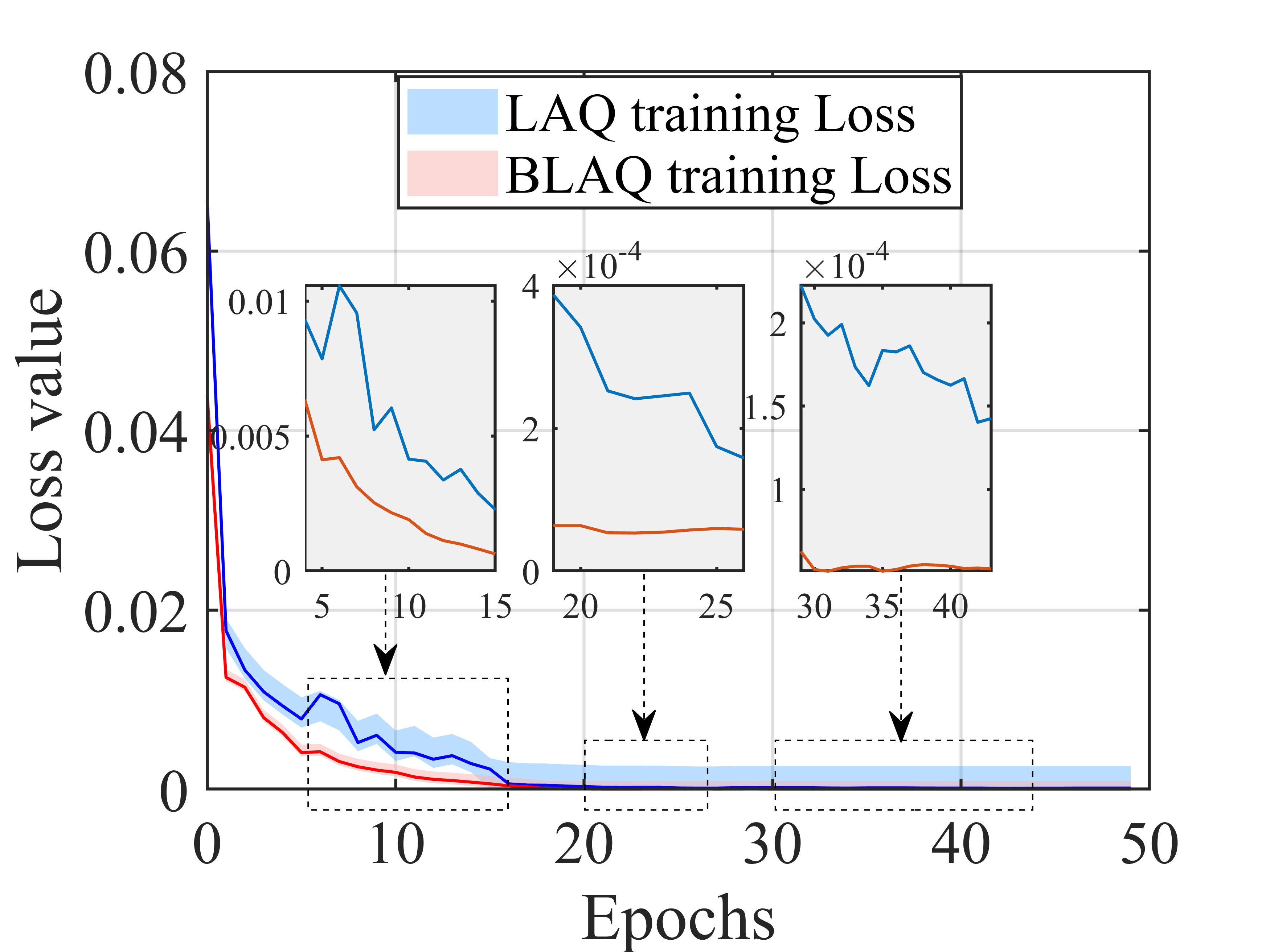}
}
\caption{Training loss trained with LAQ/baseline and BLAQ. The red curve shows the evolution of mean loss values obtained by BLAQ, while the blue one by LAQ. The shaded areas indicate the fluctuation (error) range of the mean loss curve.}
\label{4.2}
\end{figure*}

Fig. \ref{4.2} presents the training loss of different methods with the increase of the number of epochs. The results indicate that there is an obvious gap between the baseline (LAQ) and backtracking-search updating approach (BLAQ). Specifically, the training loss curve of LAQ fluctuates more drastically, and also shows slower convergence velocity than BLAQ. It may result from the quantization error from $\hat{g}^{t-1}$ in Eq. \ref{Eq3}, which is the fundamental cause of zig-zagging-like issue. By contrast, BLAQ achieves a more accurate gradient estimation during the weight training, which not only accelerates the training convergence, but also benefits the test accuracy as shown in Tables \ref{table3}-\ref{table6}. 

\subsection{Convergence Result Analysis}
\begin{figure*} \centering
\subfigure[Accuracy of CIFAR10] { \label{4.3.1}
\includegraphics[width=0.5\columnwidth]{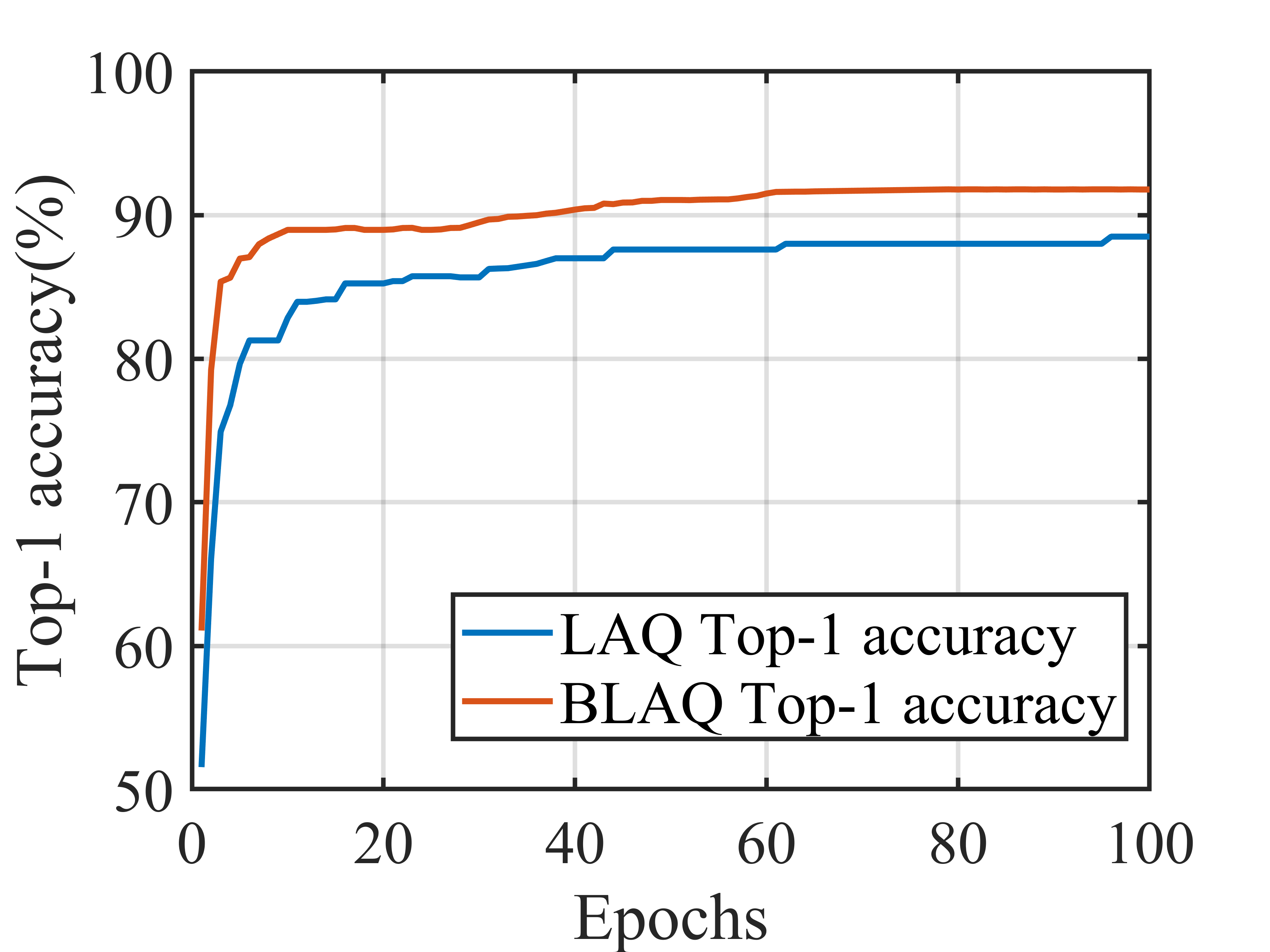}
}
\subfigure[Accuracy of MNIST] { \label{4.3.2}
\includegraphics[width=0.5\columnwidth]{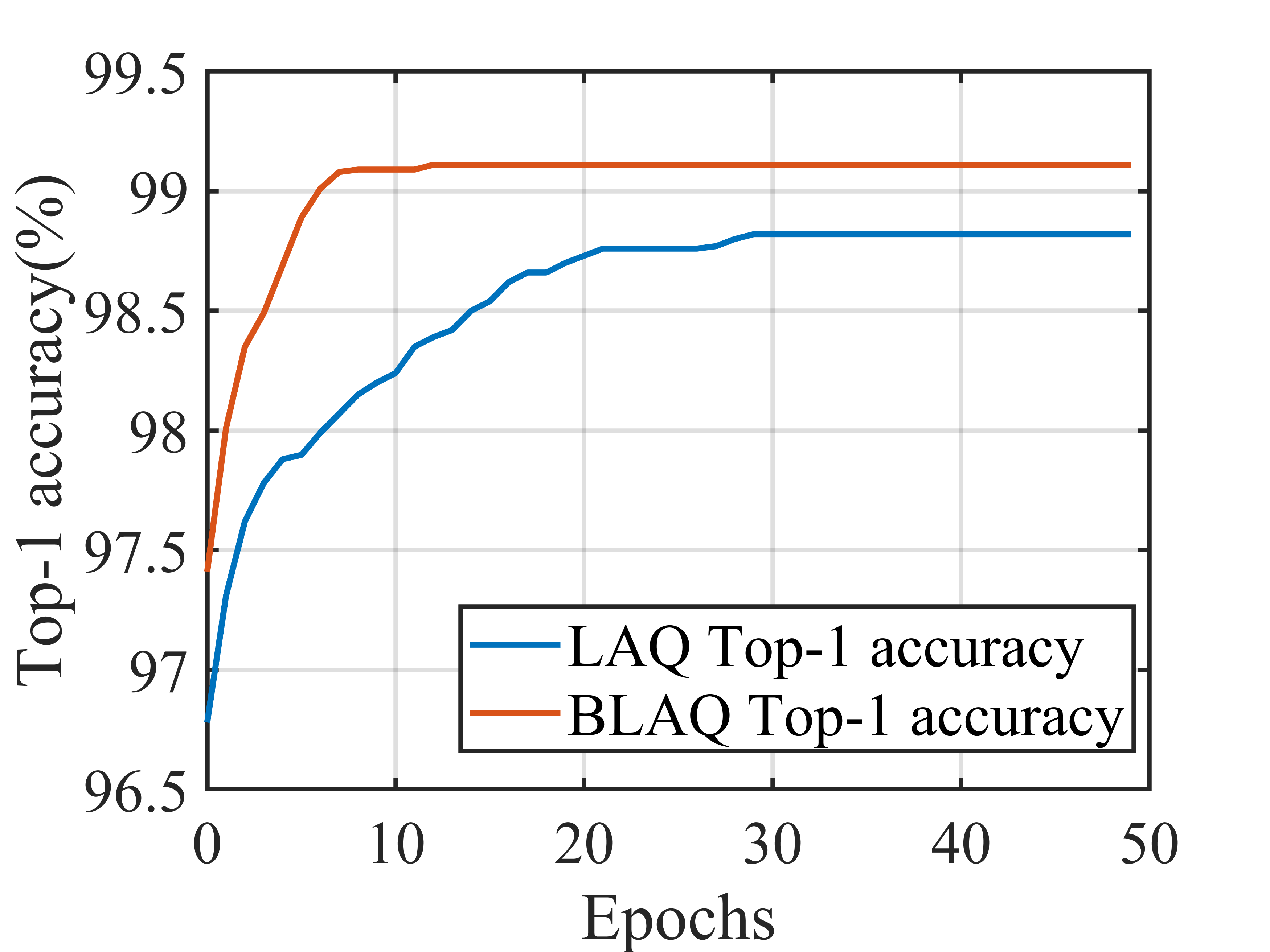}
}
\subfigure[Accuracy of SVHN] { \label{4.3.3}
\includegraphics[width=0.5\columnwidth]{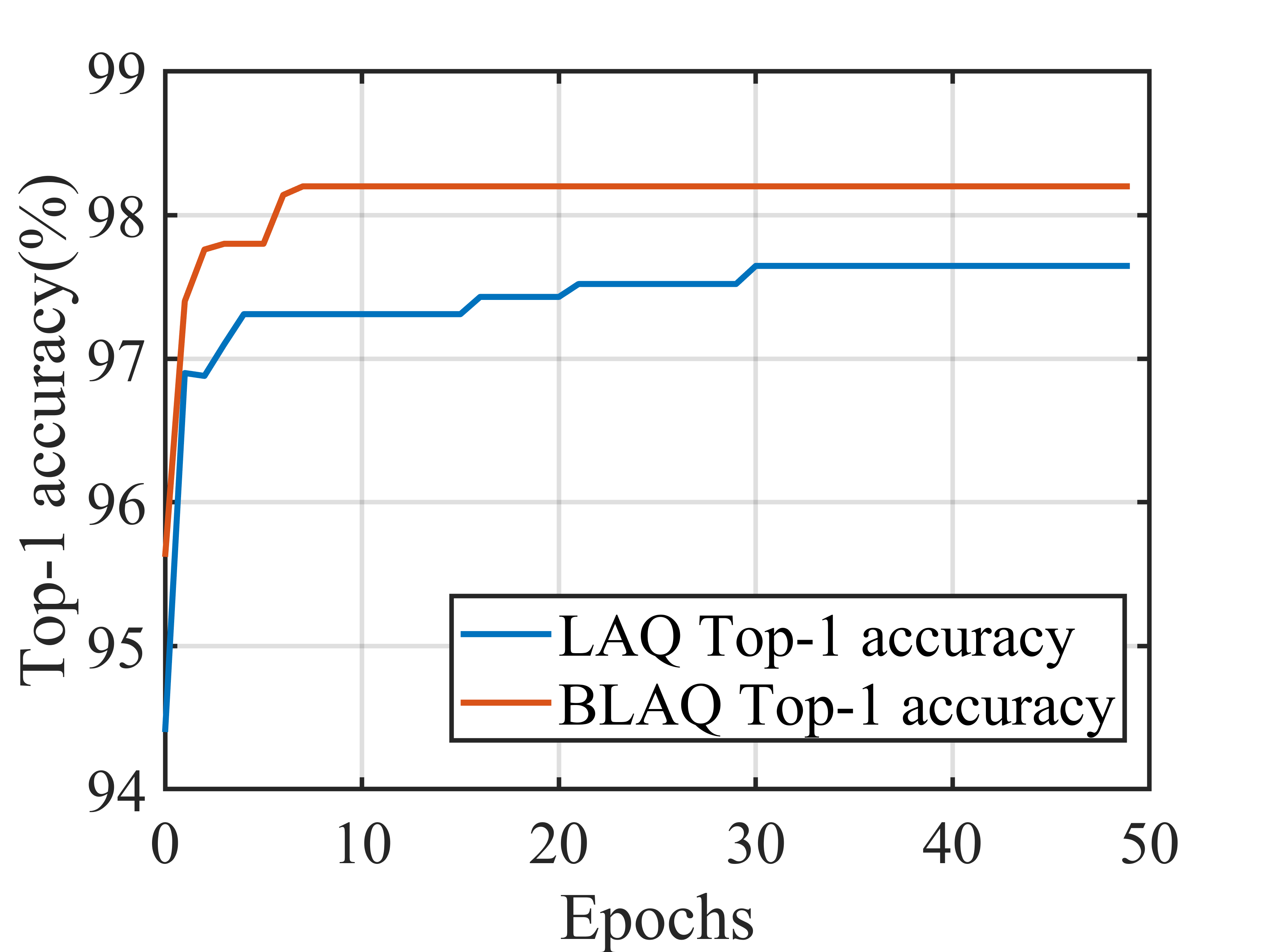}
}
\caption{Test convergence obtained by LAQ/baseline and BLAQ.}
\label{4.3}
\end{figure*}

Fig. \ref{4.3} shows the test accuracy versus training epochs on three datasets. The results show that BLAQ not only converges more stable and faster, but also gets a higher accuracy than LAQ. For example, BLAQ reaches to a high level of accuracy in few epochs on CIFAR10, but LAQ has to take more epochs to achieve similar accuracy. Similar observations can be found on MNIST and SVHN. Such encouraging results are consistent with Theorem 2. Therefore, it can be concluded that the competitive performance of BLAQ may be attributed to the backtracking-based two-stage updating, which can reduce the error caused by the gradient approximation. To ensure fair comparisons, all our comparisons are based on the results published by the original author of LAQ, while the author of LAQ didn't conduct experiments on ILSVRC12 (ImageNet). Our BLAQ method is an improvement of the LAQ method. Fig. \ref{4.2} and Fig. \ref{4.3} show a comparison of the BLAQ and LAQ training processes. Since LAQ has no experimental data on ILSVRC12 (ImageNet), Fig. \ref{4.2} and Fig. \ref{4.3} don't show the comparison of the BLAQ and LAQ training processes on ILSVRC12 (ImageNet).

\subsection{Experimental Results on CIFAR10}
In this section, we firstly compare the BLAQ with the state-of-the-art binarization baselines on CIFAR10. Table \ref{table2} has demonstrated the performance of the proposed method against the baselines on CIFAR10 in terms of Top-1 accuracy. From this table, we find that our model achieves significantly better results than these weight binarization methods (BC, BWN and LAQ). For the popular 1-bit weight quantization approaches, our method obtains at least 1.4\% improvement over BC, BWN and LAQ. Compared with the latest weight ternary method (TRQ), our method achieves better accuracy with almost 0.3\% improvement. These encouraging results validate the effectiveness of the proposed one-step forward and backtrack way in handling network binarization task.

Furthermore, we also compare our method with the adaptive multi-bit quantizers (AQ and ALQ), which assign different bitwidth for each parameter and prune these parameters according to their importance. The results in Table \ref{table2} show that our method obtains similar or even better accuracy without any expense of additional operations like pruning and multi-bit computation.
Besides, our method outperforms them with almost 0.6\% improvement, especially. The results reveal that our method without any additional operations is able to get competitive performance.
\begin{table}
	\center
	\begin{tabular}{ccc}
		\hline
		\multicolumn{2}{c}{Methods}          & Top-1 accuracy \\ \hline
		Full-precision         &32-bit           & 92.8\%        \\ 
		BC            & 1-bit           & 90.1\%        \\ 
		BWN        & 1-bit & 89.5\%        \\ 
		LAQ          & 1-bit & 89.5\%        \\ 
		TRQ*        & ternary       & 91.2\%        \\
		AQ          & multi-bit      & 90.9\%        \\ 
		ALQ			& multi-bit 			& 90.9\%    \\
		BLAQ (ours)   			& 1-bit		& \textbf{91.5}\%  \\ \hline
	\end{tabular}
	\caption{Top-1 accuracy(VGG on CIFAR10). } %The results of BWN were reported in~\cite{27r}.}
	\label{table2}
\end{table}

\subsection{Experimental Results on MNIST}
Here, the proposed method is compared with two groups of baselines on MNIST: 1) Non-pruning binarization methods (i.e., BC, BWN and LAQ); 2) Pruning-based quantization methods (i.e., ALQ, DC, and ADMM). For the first group, all methods are based on a 4-layer full-connected model; For the second group, all methods are based on a modified LeNet5 model.

\begin{table}[H]
\centering
\scalebox{0.95}{
	\begin{tabular}{ccc}\hline
	\multicolumn{2}{c}{Methods}          & Top-1 accuracy \\ \hline
			Full-precision        &32-bit           & 98.81\%        \\ 
			BC           & 1-bit           & 98.72\%        \\ 
			BWN          & 1-bit & 98.69\%        \\ 
			LAQ          & 1-bit & 98.82\%        \\ 
			BLAQ (ours)  & 1-bit		& \textbf{99.11}\%  \\ \hline
	\end{tabular}
}
\caption{Top-1 accuracy (4-layer Model on MNIST)}
	\label{table3}
\end{table}

\begin{table}[H]
\centering
\scalebox{0.95}{
	\begin{tabular}{ccc}\hline
	\multicolumn{2}{c}{Methods}          & Top-1 accuracy \\ \hline
		Full-precision        &32-bit           & 99.19\%        \\ 
		DC            & Multi-bit           & 99.26\%        \\ 
		ADMM        & Multi-bit & 99.20\%        \\ 
		ALQ          & Multi-bit & 99.12\%        \\ 
		BLAQ (ours)   			& 1-bit		& \textbf{99.38}\%  \\ \hline
		\end{tabular}
}
\caption{Top-1 accuracy  (LeNet on MNIST).}% The results of baselines were reported in \cite{34r, 27r}.}
		\label{table4}
\end{table}

Table \ref{table3} shows the comparison results in terms of Top-1 accuracy between BLAQ and the classical non-pruning binarization  baselines (BC, BWN and LAQ) on MNIST. The table shows that our method outperforms them with at least 0.29\% improvement in terms of the Top-1 accuracy, which may result from the fact BLAQ has reduced the quantization error to large extent. 

Table \ref{table4} shows the comparison results with three pruning-based binarization methods on MNIST. It can be observed that our method gains consistently best performance over the counterparts although they utilize additional compression-related operations to assist the quantization. For example, DC and ADMM use sparse tensors, which require special libraries or hardware for execution, while ALQ needs to prune networks during the multi-bit quantization. By contrast, our method is a pure binarization using no additional pruning-related operations, and more suitable to generic off-the-shelf platform. More importantly, our method achieves better results than the full-precision approaches, which again shows its promising capability in dealing with the low-bit quantization problem. 

\subsection{Experimental Results on SVHN}
In this section, we further explore the performance of our method on the well-known model (SVHN) used in LAQ \cite{26r}. 
Table \ref{table5} shows the test accuracy on SVHN. Similar to Table \ref{table3}, BLAQ performs  better than other binarization methods, and even better than the full-precision approach. For example, in comparison with BLAQ, the contrast algorithm (BWN) does not quantize the first layer and the last layer, but still gets unsatisfactory test accuracy. In addition, BLAQ achieves better performance than the approach with full-precision weights, which may be attributed to the reduction of the redundant weights in BLAQ. Overall, BLAQ also shows competitive performance on SVHN.
\begin{table}
	\center
%\scalebox{0.95}{
	\begin{tabular}{ccc}
		\hline
		\multicolumn{2}{c}{Methods}          & Top-1 accuracy \\ \hline
		Full-precision weights        &32-bit           &  97.73\%        \\ 
		BC           & 1-bit           & 97.55\%        \\ 
		BWN        & 1-bit & 97.46\%        \\ 
		LAQ         & 1-bit & 97.64\%        \\ 
		BLAQ (ours)   			& 1-bit		& \textbf{98.13}\%  \\ \hline
	\end{tabular}
%}
	\caption{Top-1 accuracy (SVHNNet on SVHN). }% The results of baselines were reported in \cite{34r, 27r}.}
	\label{table5}
\end{table}
\subsection{Experimental Results on ILSVRC12 (ImageNet)}
Note that a large batch size in large-scale datasets may influence the direction of gradient descent, as large batch size tends to help increase the model stability but decrease the generalization ability. Hence, we conduct an experiment with large batch size and quantize the weights of ResNet18 network model with low-bit (1-bit and 2-bit) on ILSVRC12 dataset, where ResNet18 is widely-used in the quantization for ILSVRC12 dataset. In this experiment, BLAQ is compared with the state-of-the-art low-bit networks.% (i.e, BWN, LR, DSQ, ALQ, TWN, LQ-Net, QIL, and OCTAV). 

Table \ref{table6} shows the Top-1 accuracy results of all methods on ILSVRC12. The results show that our method achieves the highest Top-1 accuracy, and outperforms the baselines with at least 1.13\% and 0.72\% improvements on the cases of 1-bit and  2-bit, respectively. In contrast to the popular schemes like BWN, LR, DSQ, and LQ-Net, which do not quantize the first and last layers to ensure the quantization performance since the quantization of the two layers may cause a huge accuracy degradation, the proposed BLAQ uniformly quantizes the first and last layers with 8-bit but still gets better results. In addition, we also find that the 
the recently published ALQ is not extremely low-bit in the strict sense in the quantization, where ALQ sets the bitwidth of all layers to 8-bit, which is different from our method that has physically realized 1-bit and 2-bit quantization.
The above results show that BLAQ is also able to get competitive performance in handling the quantization of deep neural networks on large-scale datasets. 
\begin{table}[t]
\centering
	\label{table6}
\scalebox{0.95}{
		\begin{tabular}{ccc}
			\hline
			\multicolumn{2}{c}{Methods}    & Top-1 accuracy \\ \hline
			Full-precision              & 32-bit         & 69.8\%        \\ 
			BWN              & 1-bit         & 60.8\%        \\ 
			LR*             & 1-bit         & 59.9\%        \\ 
			DSQ*              & 1-bit         & 63.7\%        \\ 
			ALQ              & 1.01-bit         & 65.6\%        \\ 
			BLAQ(ours)              & 1-bit         & \textbf{66.73}\%        \\  \hline
			TWN              & 2-bit         & 61.8\%        \\ 
			LR              & 2-bit         & 63.5\%        \\ 
			LQ-Net*              & 2-bit         & 68.0\%        \\ 
			QIL*              & 2-bit         & 68.1\%        \\ 
			ALQ              & 2-bit         & 68.9\%        \\ 
			OCTAV			& 4-bit 		  & 69.17\%     \\
			BLAQ(ours)              & 2-bit         & \textbf{69.62}\%        \\  \hline
	\end{tabular}
}
\caption{Top-1 accuracy (ResNet18 on ISLVRC12).}\label{table6}
\end{table}
\section{Related Work}
In the network quantization, early studies \cite{40r, 41r} use 8-bit fixed point representation, which achieves the state-of-the-art performance on ILSVRC12. Later, more efforts are made to the extremely low-bit quantization for DNNs compression \cite{40r, 41r, 42r, 43r, 44r}, which mainly discretizes activations and weights of DNNs into binary or ternary values (e.g., \{-1, +1\} or \{-1, 0, +1\}). However, simply using bit-wise operations to approximate the convolution operations of DNNs may suffer from performance deterioration. To improve the model accuracy, some quantization improvements are come up with, such as learnable quantizing\cite{45r} and non-uniform logarithmic representation \cite{46r}. In principle, previous studies treat the quantization problem as a straightforward approximation of original full-precision weights (i.e., minimizing the error between low-bit quantized weights and full-precision ones), but ignore the effect of quantization on the loss.

To overcome the above limitation, the research \cite{27r} proposes a loss-aware quantization approach, directly optimizing binarized weights to minimize the final loss. The following work \cite{26r} extends this binarization scheme to $m$-bit quantization, but Peng et al. \cite{21r} have discovered that the above approaches under certain conditions may fail to converge due to the quantization error. Zhou et al. \cite{28r} have proposed an explicit loss-aware weight quantization method by integrating the information of loss function with respect to full-precision weights into the quantization. But this method neglects the curvature information of the loss function and quantization errors, when updating the full-precision weights \cite{21r}. In this study, we discover that quantization error in LAQ may lead to severe zig-zagging-like issue and seriously slow down the model convergence, especially for extremely low bit-width quantization. Therefore, our focus is not only to compensate for gradient computation error, but also to defy the zig-zagging-like problem in the loss-aware quantization process. Consequently, our work aims to improve the iterative updating rules for weights or gradient according to the numerical stability theory, so that a new quantization framework using backtracking-based updating principle is proposed.  

\section{Conclusion} 
In this paper, we discover that the quantization error in LAQ may trigger a serious zig-zagging-like issue, which can severely slow down the model convergence. To handle the above issue, we propose a backtracking-search loss-aware quantization method for low-bit quantization. The main idea is to utilize potential information obtained from next-step exploratory search to compensate for the gradient error during the optimization. Specifically, our approach works in a one-step forward and backtrack way: At each iteration, the search first explores one forward step to find the trial gradient at the next-step, which can be adopted to assist in adjusting the gradient at current step towards the direction of fast convergence. Then we backtrack to update the quantized gradient using current gradient and trial gradient. This way is able to effectively and efficiently solve the zig-zagging-like issue. A number of theoretical analysis have validated the effectiveness of our approach in convergence. In addition, the experimental results in Tables \ref{table2}-\ref{table6} have also shown that our method performs well across different network models on three datasets, and achieves a consistent performance improvement over its counterparts. These results have demonstrated that our method is more robust than the previous quantization methods to tackle complicated network binarization scenarios with different network models.

\section{Acknowledgments}

This work is supported by the NSFC Key Supported Project of the Major Research Plan Grant (No.92267206), the National Natural Science Foundation of China (No. 62032013), the Key Technologies R\&D Program of Liaoning Province (2023JH1/10400082, 2023020456-JH/104), the National Natural Science Foundation of China (No. 62103150, 62333010), and China Postdoctoral Science Foundation (No.2021M691012).

\bibliography{aaai24}

\end{document}